\documentclass[letterpaper, 10 pt, conference]{ieeeconf}  

\IEEEoverridecommandlockouts                              

\usepackage{amsmath}
\usepackage{amssymb}
\usepackage{graphicx}
\usepackage{cite}
\usepackage{algorithm}
\usepackage{algorithmic}
\usepackage{booktabs}
\usepackage{tikz}
\usetikzlibrary{arrows.meta,positioning,fit,backgrounds,calc}
\usepackage{xcolor}
\usepackage{hyperref}
\usepackage{balance}

\definecolor{phaseblue}{RGB}{210,228,248}
\definecolor{phaseorange}{RGB}{255,235,210}
\definecolor{boxblue}{RGB}{155,195,235}
\definecolor{boxgreen}{RGB}{160,210,170}
\definecolor{boxorange}{RGB}{245,195,130}
\definecolor{boxpurple}{RGB}{205,180,230}
\definecolor{boxred}{RGB}{235,170,170}
\definecolor{boxgray}{RGB}{220,220,220}
\definecolor{darktext}{RGB}{50,50,50}

\title{\LARGE \bf
Projected Gradient Unlearning for Text-to-Image Diffusion Models: Defending Against Concept Revival Attacks
}

\author{Aljalila Aladawi$^{1}$, Mohammed Talha Alam$^{1}$, and Fakhri Karray$^{1,2}$%
\thanks{$^{1}$Mohamed bin Zayed University of Artificial Intelligence, Abu Dhabi, UAE.
        {\tt\small \{aljalila.aladawi, mohammed.alam, fakhri.karray\}@mbzuai.ac.ae}}%
\thanks{$^{2}$University of Waterloo, Ontario, Canada.}%
\thanks{Appendix available
        \href{https://drive.google.com/file/d/1sMin218Tm79RTh60nMNY56ydJeFX5yXo/view?usp=sharing}{\textit{here}}. Code available at: \href{https://github.com/Aj1aj2/PGU_stable_diffusion}{\textit{GitHub}}}
}

\begin{document}

\maketitle
\thispagestyle{empty}
\pagestyle{empty}

\begin{abstract}

Machine unlearning for text-to-image diffusion models aims to selectively remove undesirable concepts from pre-trained models without costly retraining. Current unlearning methods share a common weakness: erased concepts return when the model is fine-tuned on downstream data, even when that data is entirely unrelated. We adapt Projected Gradient Unlearning (PGU) from classification to the diffusion domain as a post-hoc hardening step. By constructing a Core Gradient Space (CGS) from the retain concept activations and projecting gradient updates into its orthogonal complement, PGU ensures that subsequent fine-tuning cannot undo the achieved erasure. Applied on top of existing methods (ESD, UCE, Receler), the approach eliminates revival for style concepts and substantially delays it for object concepts, running in roughly 6 minutes versus the $\sim$2 hours required by Meta-Unlearning. PGU and Meta-Unlearning turn out to be complementary: which performs better depends on how the concept is encoded, and retain concept selection should follow visual feature similarity rather than semantic grouping.

\end{abstract}

\section{INTRODUCTION}

\begin{figure}[t]
\centering
\def\IW{3.25cm}
\def\IH{3.25cm}
\def\XSEP{3.53}
\def\XC{3.40}
\def\XD{3.58}
\def\XM{3.55}
\def\RY{-4.35}

\begin{tikzpicture}[x=1cm,y=1cm,font=\scriptsize,>=Stealth]

  \node[anchor=north west, inner sep=0] (A) at (0,0)
    {\includegraphics[width=\IW,height=\IH]{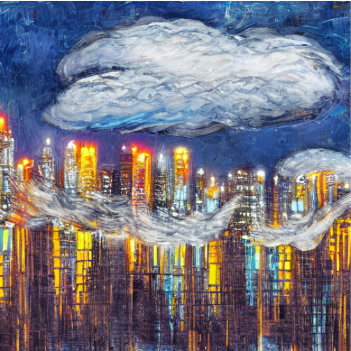}};
  \node[anchor=north west, inner sep=0] (B) at (\XSEP,0)
    {\includegraphics[width=\IW,height=\IH]{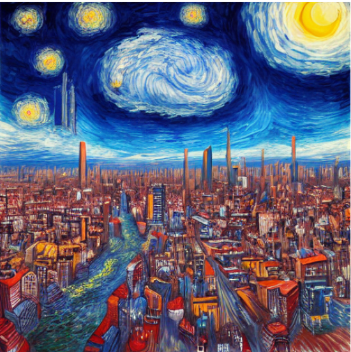}};

  \node[font=\scriptsize\bfseries, text=red!70] at (\XM,0.34) {Without PGU};

  \node[anchor=south, text=black!60, font=\tiny\bfseries] at (1.63,0.03) {Before};
  \node[anchor=south, text=black!60, font=\tiny\bfseries] at (5.16,0.03) {After};

  \draw[-{Stealth[length=3.5pt]}, thick, red!70] (\XC,-1.63) -- (\XD,-1.63);
  \node[text=red!70, font=\tiny\bfseries] at (\XM,-1.15) {Fine-tune};

  \node[anchor=north, fill=red!80!black, text=white,
        font=\bfseries\tiny, rounded corners=1pt, inner sep=1.2pt]
    at (5.16,-3.31) {Concept Revived};

  \draw[dashed, black!20] (-0.05,-3.78) -- (6.95,-3.78);

  \node[anchor=north west, inner sep=0] (C) at (0,\RY)
    {\includegraphics[width=\IW,height=\IH]{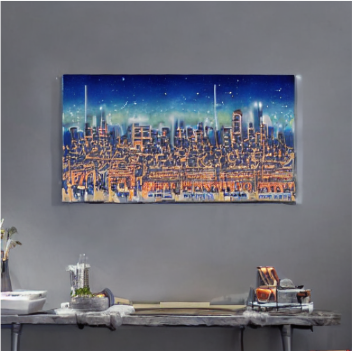}};
  \node[anchor=north west, inner sep=0] (D) at (\XSEP,\RY)
    {\includegraphics[width=\IW,height=\IH]{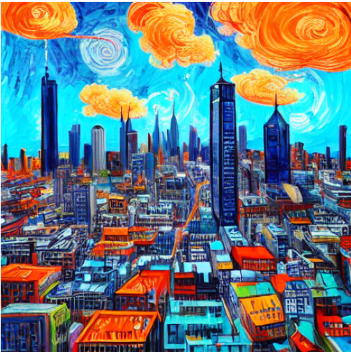}};

  \node[font=\scriptsize\bfseries, text=green!50!black] at (\XM,\RY+0.34) {With PGU};

  \node[anchor=south, text=black!60, font=\tiny\bfseries] at (1.63,\RY+0.03) {Before};
  \node[anchor=south, text=black!60, font=\tiny\bfseries] at (5.16,\RY+0.03) {After};

  \draw[-{Stealth[length=3.5pt]}, thick, green!50!black] (\XC,\RY-1.63) -- (\XD,\RY-1.63);
  \node[text=green!50!black, font=\tiny\bfseries] at (\XM,\RY-1.15) {Fine-tune};

  \node[anchor=north, fill=green!50!black, text=white,
        font=\bfseries\tiny, rounded corners=1pt, inner sep=1.2pt]
    at (5.16,\RY-3.31) {Erasure Preserved};

  \node[text=black!50, font=\tiny\bfseries\itshape, text width=6.8cm, align=center]
    at (\XM,\RY-3.72)
    {Prompt: ``a city skyline in Van Gogh style'' --- style erased by ESD-U};

\draw[dashed, black!20] (3.4,0.42) -- (3.4,\RY-3.45);

\end{tikzpicture}

\vspace{-2mm}
\caption{Fine-tuning vulnerability and PGU defense. \textit{Top:} without PGU, the Van Gogh style revives after fine-tuning on unrelated data. \textit{Bottom:} with PGU hardening, erasure is preserved during downstream adaptation.}
\label{fig:teaser}
\end{figure}

Diffusion models such as Stable Diffusion, DALL$\cdot$E, and Midjourney enable high-quality text-to-image synthesis, driving widespread adoption across creative and scientific domains~\cite{paper1,alam2024flare,alam2024introducing,paper8,paper30}. Training on massive web-scraped datasets~\cite{schuhmann2022laion} raises serious ethical and legal concerns: models may memorize private, copyrighted, or explicit content. The EU GDPR~\cite{gdpr2018,hoofnagle2019gdpr} and CCPA~\cite{delatorre2018ccpa} both grant a ``right to be forgotten'', which translates directly into requirements for selective content removal from deployed models. Full retraining is rarely feasible at scale, so machine unlearning~\cite{cao2015unlearning,ginart2019deletion,golatkar2020eternal} has become the standard approach for selective concept removal. Despite significant progress~\cite{paper9,paper35,paper36,paper24,fan2024salun}, a fundamental flaw persists: \textit{unlearned concepts revive when models undergo subsequent fine-tuning, even on unrelated data}~\cite{paper1,qi2023finetuning}, threatening legal compliance for any model subject to downstream adaptation~\cite{paper30} (Fig.~\ref{fig:teaser}).

Recent benchmarking work confirms that robustness to benign fine-tuning is largely under-evaluated in current safety and unlearning methods~\cite{alam2025spqr}, and none of the existing approaches explicitly enforce it. Retained concepts remain semantically linked to erased ones, providing pathways through which erasure can be undone~\cite{paper30}. We therefore look for a defense that hardens already-erased models without modifying their training procedures. Our approach adapts Projected Gradient Unlearning (PGU)~\cite{paper5} to the diffusion domain and applies it as a post-hoc step: gradient updates are projected onto a subspace spanned by a small set of reference concepts. We refer to these as \textit{retain concepts}: text prompts chosen for their visual proximity to the erased concept, distinct from the full retaining dataset used in the original classification-domain PGU~\cite{paper5}. The core intuition is geometric: if retain concepts share visual features with the erased concept,  their U-Net activations occupy overlapping regions of feature space. A Core Gradient Space (CGS) built from such retained concepts, therefore, covers the gradient directions relevant to the generation of the erased concept. Projecting fine-tuning gradients onto this CGS prevents parameter updates most likely to restore the erased concept's visual features, regardless of the data used for fine-tuning.

\textbf{Contributions:} (1) First adaptation of PGU to diffusion models as a post-hoc hardening step, with novel loss design, U-Net activation-based SVD, and hyperparameter recalibration.
(2) Efficient modular defense compatible with any base unlearning method. (3) Validation across three operational conditions that cover vulnerable, robust, and failed-erasure cases. (4) Empirical evidence that visual feature similarity is the key factor for retain concept selection.

\section{Background}
 \begin{figure*}[t]
\centering
\resizebox{\textwidth}{!}{%
\begin{tikzpicture}[x=1cm,y=1cm,font=\small,
  IO/.style   ={rectangle,rounded corners=2pt,draw=black!50,fill=boxgray,
                minimum width=2.1cm,minimum height=0.7cm,
                align=center,inner sep=3pt,text width=2.0cm},
  PROC/.style ={rectangle,rounded corners=3pt,draw=#1!80!black,fill=#1!35,
                minimum width=2.4cm,minimum height=0.7cm,
                align=center,inner sep=3pt,text width=2.3cm},
  MATH/.style ={rectangle,rounded corners=2pt,draw=#1!80!black,fill=#1!20,
                minimum width=2.6cm,minimum height=0.8cm,
                align=center,inner sep=3pt,text width=2.5cm,font=\scriptsize},
  RES/.style  ={rectangle,rounded corners=3pt,draw=#1!80!black,fill=#1!30,
                minimum width=2.4cm,minimum height=0.8cm,
                align=center,inner sep=3pt,text width=2.3cm},
  SMALL/.style={rectangle,rounded corners=2pt,draw=black!45,fill=boxgray!70,
                minimum width=1.5cm,minimum height=0.5cm,
                align=center,inner sep=2pt,text width=1.4cm,font=\scriptsize},
  GUAR/.style ={rectangle,rounded corners=3pt,draw=green!55!black,fill=green!12,
                minimum width=3.0cm,minimum height=2.6cm,
                align=center,inner sep=5pt,text width=2.9cm,font=\scriptsize},
  LEG/.style  ={rectangle,rounded corners=2pt,draw=black!30,fill=white,
                inner sep=5pt,font=\scriptsize,align=left},
  arr/.style  ={-{Stealth[length=5pt,width=4pt]},thick,draw=black!55},
  darr/.style ={-{Stealth[length=5pt,width=4pt]},thick,draw=#1,dashed},
  lbl/.style  ={font=\scriptsize,text=black!55},
]
 
\def\PY{2.5}
 
\node[IO]          (P1A) at (1.5,  \PY)      {Retain\\Prompts $c_r$};
\node[PROC=boxblue](P1B) at (4.3,  \PY)      {CLIP Text\\Encoder $\tau$};
\node[IO]          (P1N) at (4.3,  \PY-1.25) {Noise\\$z\!\sim\!\mathcal{N}(0,I)$\\$t\!\sim\!\mathcal{U}[0,T]$};
\node[PROC=boxgreen](P1C) at (7.4, \PY)
  {U-Net Forward\\$\epsilon_\theta(z_t,t,\tau(c_r))$\\{\scriptsize hooks collect $r^l$}};
\node[SMALL]       (RMH) at (7.4,  \PY-1.25) {Remove\\Hooks};
\node[MATH=boxpurple](P1D) at (10.6,\PY)
  {Covariance\\$C^l \mathrel{+}= (r^l)^\top r^l$\\{\scriptsize $O(d^2)$ memory}};
\node[MATH=boxorange](P1E) at (13.8,\PY)
  {Eigendecomp.\\$C^l = U^l\Lambda^l(U^l)^\top$\\
   {\scriptsize select $k_l$: $\textstyle\sum_{i=1}^{k_l}\!\sigma_i\geq\gamma\sum_i\sigma_i$}};
\node[RES=boxred]  (P1F) at (17.2, \PY)
  {Projection\\Matrices $P^l$\\$P^l\!=\!M^l(M^l)^\top$\\{\scriptsize $M^l=U^l_{:,1:k_l}$}};
 
\draw[arr] (P1A)--(P1B);
\draw[arr] (P1B)--(P1C) node[lbl,above,midway]{$\tau(c_r)$};
\draw[arr] (P1N.east) -- ++(0.5,0) |- (P1C.west);
\draw[arr] (P1C)--(P1D);
\draw[arr] (P1D)--(P1E);
\draw[arr] (P1E)--(P1F);
\draw[arr,draw=black!40] (P1C.south)--(RMH.north);
 
\begin{scope}[on background layer]
  \node[rectangle,rounded corners=4pt,fill=phaseblue,fill opacity=0.4,
        inner sep=10pt,
        fit=(P1A)(P1B)(P1N)(P1C)(RMH)(P1D)(P1E)(P1F)](Ph1box){};
\end{scope}
\node[font=\normalsize\bfseries,text=blue!65!black, anchor=south west]
  at (-1.0, 4.0)
  {Phase 1 --- Retain Subspace Construction
   \textnormal{\scriptsize(one-time; hooks registered, used, then removed)}};
 
\def\QY{-2.0}
 
\node[IO](P2Af) at (1.5,-1.1) {Forget $c_f$\\{\scriptsize (ESD)}};
\node[IO](P2Ar) at (1.5,-2.9) {Retain $c_r$\\{\scriptsize (UCE/Rec.)}};
 
\node[PROC=boxblue](P2B) at (4.3,\QY) {CLIP Text\\Encoder $\tau$};
\draw[arr] (P2Af.east) -- (3.1,-1.1) -- (3.1,-1.65) -- (P2B.north);
\draw[arr] (P2Ar.east) -- (3.1,-2.9) -- (3.1,-2.35) -- (P2B.south);
 
\node[IO](P2N) at (4.3,\QY-1.4)
  {Noise\\$z\!\sim\!\mathcal{N}(0,I)$\\$t\!\sim\!\mathcal{U}[0,T]$};
 
\node[PROC=boxgreen](P2C) at (7.4,\QY)
  {U-Net Forward\\{\scriptsize ESD: $\epsilon_\theta(z_t,t,\tau(c_f))$}\\
   {\scriptsize UCE/Rec.: $\epsilon_\theta(z_t,t,\tau(c_r))$}};
\draw[arr] (P2B)--(P2C);
\draw[arr] (P2N.east) -- ++(0.5,0) |- (P2C.west);
 
\node[MATH=boxorange](P2D) at (10.6,\QY)
  {Loss $\mathcal{L}$\\
   {\scriptsize ESD: $\|\epsilon_\theta\!-\!\epsilon_\text{target}\|^2$}\\
   {\scriptsize UCE/Rec.: $\mathcal{L}_\text{retain}$}};
\draw[arr] (P2C)--(P2D);
 
\node[PROC=boxpurple](P2E) at (13.8,\QY)
  {Backward Pass\\$\nabla_{w^l}\mathcal{L}$};
\draw[arr] (P2D)--(P2E);
 
\node[MATH=boxred](P2F) at (17.2,\QY)
  {Gradient Projection\\
   $\nabla_{w^l}\!\mathcal{L}_\perp=$\\
   $\nabla_{w^l}\!\mathcal{L}-\nabla_{w^l}\!\mathcal{L}\!\cdot\! P^l$};
\draw[arr] (P2E)--(P2F);
 
\node[RES=boxgreen](P2G) at (20.4,\QY)
  {Optimizer Step\\$\theta\!\leftarrow\!\theta - \alpha\nabla_\theta\mathcal{L}_\perp$};
\draw[arr] (P2F)--(P2G);
 
\draw[darr=red!65]
  (P1F.south) -- (P2F.north)
  node[lbl,midway,right=3pt]{\scriptsize $P^l$ (cached)};
 
\begin{scope}[on background layer]
  \node[rectangle,rounded corners=4pt,fill=phaseorange,fill opacity=0.4,
        inner sep=10pt,
        fit=(P2Af)(P2Ar)(P2B)(P2N)(P2C)(P2D)(P2E)(P2F)(P2G)](Ph2box){};
\end{scope}
\node[font=\normalsize\bfseries,text=orange!75!black, anchor=north west]
  at (-1.0,-4.5)
  {Phase 2 --- Gradient-Projected Hardening
   \textnormal{\scriptsize(repeated for $N$ training steps)}};
 
\node[GUAR] (GUAR) at (26.0, 0.25)
  {\textbf{Orthogonality}\\
   \textbf{Guarantee}\\[4pt]
   $\nabla_{w^l}\mathcal{L}_\perp \perp \mathrm{CGS}$\\[2pt]
   $\Rightarrow w'r=wr$\\
   $\forall r\in\mathrm{CGS}$\\[4pt]
   Retain outputs\\
   preserved exactly.\\[2pt]
   Fine-tuning cannot\\
   undo erasure.};
 
\draw[arr,draw=green!55!black]
  (18.4, 2.5) -- (23.5, 2.5) -- (23.5, 1.55) -- (24.5, 1.55);
 
\draw[arr,draw=green!55!black]
  (21.6,-2.0) -- (23.5,-2.0) -- (23.5,-1.05) -- (24.5,-1.05);
 
\node[LEG] at (10.6, 0.25)
  {\textbf{Gradient Space Partition}\\[3pt]
   {\color{blue!65!black}\rule{7pt}{7pt}}~\textbf{CGS}: Core Gradient Space\\
   \quad retain directions --- blocked by $P^l$\\[2pt]
   {\color{green!55!black}\rule{7pt}{7pt}}~\textbf{RGS}: Residual Gradient Space\\
   \quad safe directions for unlearning\\[3pt]
   $\gamma\in[0.5,0.7]$: CGS variance fraction};
 
\node[IO](EM) at (-1.0,0.25) {Erased\\Model $\theta$};
 
\draw[arr] (EM.north) -- (-1.0,1.4) -- (1.5,1.4) -- (P1A.south);
 
\draw[arr] (EM.south) -- (-1.0,-1.1) -- (P2Af.west);
 
\draw[arr] (EM.south) -- (-1.0,-2.9) -- (P2Ar.west);
 
\end{tikzpicture}
}
\caption{Internal workflow of PGU adapted for diffusion models.
\textit{Phase~1 (blue, run once):} retain concept prompts $c_r$ are CLIP-encoded and fed through the U-Net; forward hooks collect per-layer input activations $r^l$, which are then removed. Activations accumulate into covariance matrices $C^l$ eigendecomposed as $C^l=U^l\Lambda^l(U^l)^\top$ to identify the Core Gradient Space (CGS); projection matrices $P^l$ are cached.
\textit{Phase~2 (orange, per step):} for ESD-erased models the forget concept $c_f$ drives the ESD loss; for UCE/Receler the retain concept $c_r$ drives the standard denoising loss. Each gradient is projected orthogonal to the CGS before the optimizer step, guaranteeing subsequent fine-tuning cannot undo the achieved erasure.}
\label{fig:architecture}
\end{figure*}
 
 
\subsection{Machine Unlearning for Diffusion Models}

An effective unlearning method must satisfy~\cite{paper24}: \textit{efficacy}
(target not generated), \textit{generality} (synonyms also erased), and
\textit{specificity} (unrelated concepts intact). \textbf{ESD}~\cite{paper9}
uses the model's own score function to steer generation away from a concept,
with ESD-U modifying all non-cross-attention layers for broad suppression and
ESD-X modifying only cross-attention for surgical erasure.
\textbf{UCE}~\cite{paper35} computes closed-form cross-attention weight updates
with explicit preservation terms. \textbf{Receler}~\cite{paper36} inserts
lightweight adapters trained with adversarial prompt learning. Additional methods include MACE~\cite{paper24}, SalUn~\cite{fan2024salun} and SPM~\cite{lyu2024spm}.
 
\subsection{The Fine-Tuning Vulnerability}
 
George et al.~\cite{paper1} show that fine-tuning an unlearned model on arbitrary concepts consistently revives erased content across all examined methods. Their attack employs a curriculum strategy, sequentially fine-tuning on data in ascending semantic distance from a target; the \textit{revival point} is the earliest 
checkpoint where revival is detected.
 
\subsection{Meta-Unlearning and Projected Gradient Unlearning}
 
Meta-Unlearning~\cite{paper30} addresses the vulnerability by simulating attacks during training via bi-level optimization, inducing self-destruction of related benign concepts when malicious fine-tuning is attempted, at a substantial memory and compute cost. In its original classification setting, PGU~\cite{paper5} partitions gradient space into a Core Gradient Space (CGS) containing directions important for retained knowledge and a Residual Gradient Space (RGS) for safe updates, projecting unlearning gradients onto the RGS to preserve retained knowledge while performing unlearning. Originally developed as an unlearning method with discrete labels and feedforward architectures, PGU requires adaptation in both role and form for diffusion models: we adapt its gradient projection mechanism as a post-hoc defense against fine-tuning-based revival.


\section{Proposed Methodology}
\subsection{Geometric Motivation}
 
When a fine-tuning gradient satisfies $\nabla_\theta\mathcal{L}_\text{FT}^\top
\nabla_\theta\mathcal{L}_\text{unlearn} > 0$, it simultaneously reduces the fine-tuning loss and increases the unlearning loss, undoing erasure~\cite{paper1}. 
If retain concepts are chosen to be \textit{visually similar} to the erased concept, their U-Net activations span gradient directions relevant to that concept's generation; projecting updates onto a subspace orthogonal to these directions blocks the pathways most likely to restore erased visual features.
 
\subsection{Adapting PGU to Diffusion Models}
 
 

Translating this geometric intuition into a working algorithm requires three changes to the original classification-domain PGU: a new loss formulation suited to denoising objectives, SVD computed over U-Net activations rather than layer outputs, and a recalibrated $\gamma$ range.

\textbf{Loss function.} For ESD-erased models, we continue the ESD guidance using frozen teacher parameters $\theta^*$:
\begin{equation}
\boldsymbol{\epsilon}_\text{target} = \boldsymbol{\epsilon}_{\theta^*}(z_t,t,\varnothing)
- \eta\!\left[\boldsymbol{\epsilon}_{\theta^*}(z_t,t,c_f)
- \boldsymbol{\epsilon}_{\theta^*}(z_t,t,\varnothing)\right]
\end{equation}
For closed-form and adapter methods (UCE, Receler), the standard denoising loss on retain prompts $c_r$ is used: $\mathcal{L}_\text{retain} =
\mathbb{E}_{z,\epsilon,t}\|\epsilon_\theta(z_t,t,c_r) - \epsilon\|^2$.
 
\textbf{SVD over U-Net activations.} Forward hooks are registered on all trainable linear and convolutional U-Net layers, capturing activations during denoising conditioned on retain prompts, which is the space where SGD updates operate~\cite{paper5, zhang2017understanding}. Covariance is accumulated as $C^l = \sum_i(r^l_i)^\top r^l_i$, reducing memory
from $O(nd)$ to $O(d^2)$ per layer.
 
\textbf{Gamma re-calibration.} The classification default $\gamma\!\in\![0.9,0.95]$ 
is too aggressive for diffusion U-Nets, where extensive feature sharing through 
skip connections and cross-attention distributes activations more broadly across 
the eigenspectrum. We find $\gamma\!\in\![0.5,0.7]$ optimal for style concepts 
and $\gamma\!\in\![0.7,0.9]$ for object concepts; $\gamma\!=\!0.7$ serves as a 
reliable universal default when concept type is unknown.
 
\subsection{Algorithm}
 
PGU operates as a two-phase post-hoc step on any already-erased SD model 
(Fig.~\ref{fig:architecture}, Algorithm~\ref{alg:pgu}). Phase~1 precomputes per-layer projection matrices. Phase~2 applies gradient-projected hardening, adding only a matrix multiplication per step. Orthogonality guarantees $w'r = wr$ for any retain-subspace input~\cite{paper5}, preserving outputs on retained concepts exactly.
 
\begin{algorithm}[H]
\caption{PGU Hardening for Diffusion Models}
\label{alg:pgu}
\footnotesize
\begin{algorithmic}[1]
\REQUIRE Erased model $\theta$, forget $c_f$, retain $\{c_r^i\}$,
$\gamma$, lr $\alpha$, steps $N$
\STATE \textbf{Phase 1 -- Construct Retain Subspace}
\STATE Register hooks on all trainable layers; init $C^l\!\leftarrow\!\mathbf{0}$
\FOR{each retain prompt $p$}
  \STATE Sample $z\!\sim\!\mathcal{N}(0,I)$, $t\!\sim\!\mathcal{U}[0,T]$; forward pass
  \STATE $C^l \leftarrow C^l + (r^l)^\top r^l$ for each $l$
\ENDFOR
\FOR{each layer $l$}
  \STATE Eigendecompose $C^l$; select $k_l$ s.t.\ $\textstyle\sum_{i=1}^{k_l}\!\sigma_i\geq\gamma\textstyle\sum_i\sigma_i$
  \STATE $P^l \leftarrow M^l(M^l)^\top$, where $M^l = [u_1,\ldots,u_{k_l}]$
\ENDFOR; remove hooks
\STATE \textbf{Phase 2 -- Gradient-Projected Hardening}
\FOR{step $= 1$ to $N$}
  \STATE Compute $\mathcal{L}$ (Eq.~1 for ESD, else $\mathcal{L}_\text{retain}$); backward
  \FOR{each layer $l$}
    \STATE $\nabla_{w^l}\mathcal{L} \leftarrow \nabla_{w^l}\mathcal{L} - \nabla_{w^l}\mathcal{L}\cdot P^l$
  \ENDFOR
  \STATE $\theta \leftarrow \theta - \alpha\,\nabla_\theta\mathcal{L}_\perp$
\ENDFOR
\RETURN Hardened model $\theta$
\end{algorithmic}
\end{algorithm}
 
\subsection{Comparison with Meta-Unlearning}
 
Meta-Unlearning~\cite{paper30} is \textit{anticipatory}: it simulates specific attack patterns via bi-level optimization, requiring nested second-order gradients and inducing self-destruction that cannot distinguish malicious from legitimate adaptation. PGU is \textit{mechanistic}: it prevents the geometric condition enabling revival, makes no assumptions about future fine-tuning form, and preserves utility in directions orthogonal to the CGS. Practically, PGU requires ${\sim}$6 minutes on consumer hardware (${\sim}$12 GB VRAM) versus ${\sim}$2 hours on enterprise hardware ($>$40 GB VRAM) for Meta-Unlearning.


\section{Experimental Setup}
The experiments use SD v1.4 on an NVIDIA A100 (40 GB VRAM). PGU hardening uses 100 training steps at lr $10^{-5}$ with 100 retain samples. Fine-tuning images are generated by SD3 Medium to prevent data contamination~\cite{paper1}. Four unlearning methods are evaluated: ESD-U and ESD-X~\cite{paper9}, UCE~\cite{paper35}, and Receler~\cite{paper36}. Two concepts are studied: \textit{Van Gogh} exemplifying distributed style encoding via multi-layer feature correlations~\cite{paper1,7780634}, and \textit{Golf Ball} exemplifying compact object encoding concentrated in higher-level representations. Each is evaluated using a 10-stage curriculum, progressing in ascending order of semantic distance from the target. Revival is detected by a dual threshold: The CLIP score threshold set at 0.02 below the 
original SD v1.4 model's score and the ViT classifier accuracy threshold set at $\geq$30\%. The \textit{revival
point} $C_k$ is the first checkpoint where both are exceeded simultaneously. The configurations fall into three test cases: \textit{Case 1 (Vulnerable)}: erasure succeeds, but revival occurs; \textit{Case 2 (Robust)}: erasure persists without defense, and \textit{Case 3 (Failed)}: initial unlearning is unsuccessful. This framework validates three PGU properties: improvement for vulnerable models, preservation for robust ones, and appropriate failure for inadequate erasure.

\balance
\section{Results and Analysis}
\subsection{Baseline Vulnerability Assessment}

Table~\ref{tab:baseline} and Fig.~\ref{fig:all_methods} establish baseline profiles across all methods and concepts. Among vulnerable configurations, ESD-U Golf Ball (C1) represents the most severe case, reviving with only 50 fine-tuning images. The difference in revival timing between the style (C6) and the object (C1) can be traced to how each concept is stored in the network. Artistic style is encoded through correlations across multiple network components~\cite{7780634}, and recovering it requires reactivating many of them simultaneously, which early fine-tuning does not accumulate enough gradient signal to achieve. Object content is more concentrated, so fewer parameters need to shift, and the concept returns with far less data.

The difference between ESD-U (C6) and ESD-X (no revival) for Van Gogh comes down to which parameters each method modifies. ESD-U targets non-cross-attention weights throughout the U-Net, which are the same weights updated during standard fine-tuning, leaving direct gradient pathways open to reverse the erasure.ESD-X targets only cross-attention, which Gandikota et al.~\cite{paper9} describe as ``a gateway to the prompt''. Restoring the precise text-visual association disrupted by ESD-X requires more concentrated gradient information than reversing the broadly distributed ESD-U changes, explaining the $4\times$ difference in baseline robustness. Receler achieves the strongest baseline across both concepts~\cite{paper1}. UCE Golf Ball fails initial erasure entirely (80\% pre-attack), likely due to sensitivity to hyperparameter configuration and text
embedding formats.
\begin{table}[H]
\centering
\caption{Baseline finetuning results and test case classification.}
\label{tab:baseline}
\footnotesize
\setlength{\tabcolsep}{4pt}
\begin{tabular}{@{}llccc@{}}
\toprule
\textbf{Method} & \textbf{Concept} & \textbf{Pre-Atk} & \textbf{Revival} & \textbf{Case} \\
\midrule
ESD-U   & Van Gogh  & $\sim$5\% & C6   & Vulnerable \\
ESD-U   & Golf Ball & $\sim$5\% & C1   & Vulnerable \\
ESD-X   & Van Gogh  & $\sim$5\% & None & Robust \\
ESD-X   & Golf Ball & $\sim$5\% & C4   & Vulnerable \\
UCE     & Van Gogh  & 11\%      & None & Robust \\
UCE     & Golf Ball & 80\%      & N/A  & Failed \\
Receler & Van Gogh  & 0\%       & None & Robust \\
Receler & Golf Ball & $\sim$5\% & None & Robust \\
\bottomrule
\end{tabular}
\end{table}
\subsection{Cases 1 and 2: Vulnerable and Robust Models}
\begin{figure}[H]
\centering
\includegraphics[width=\columnwidth]{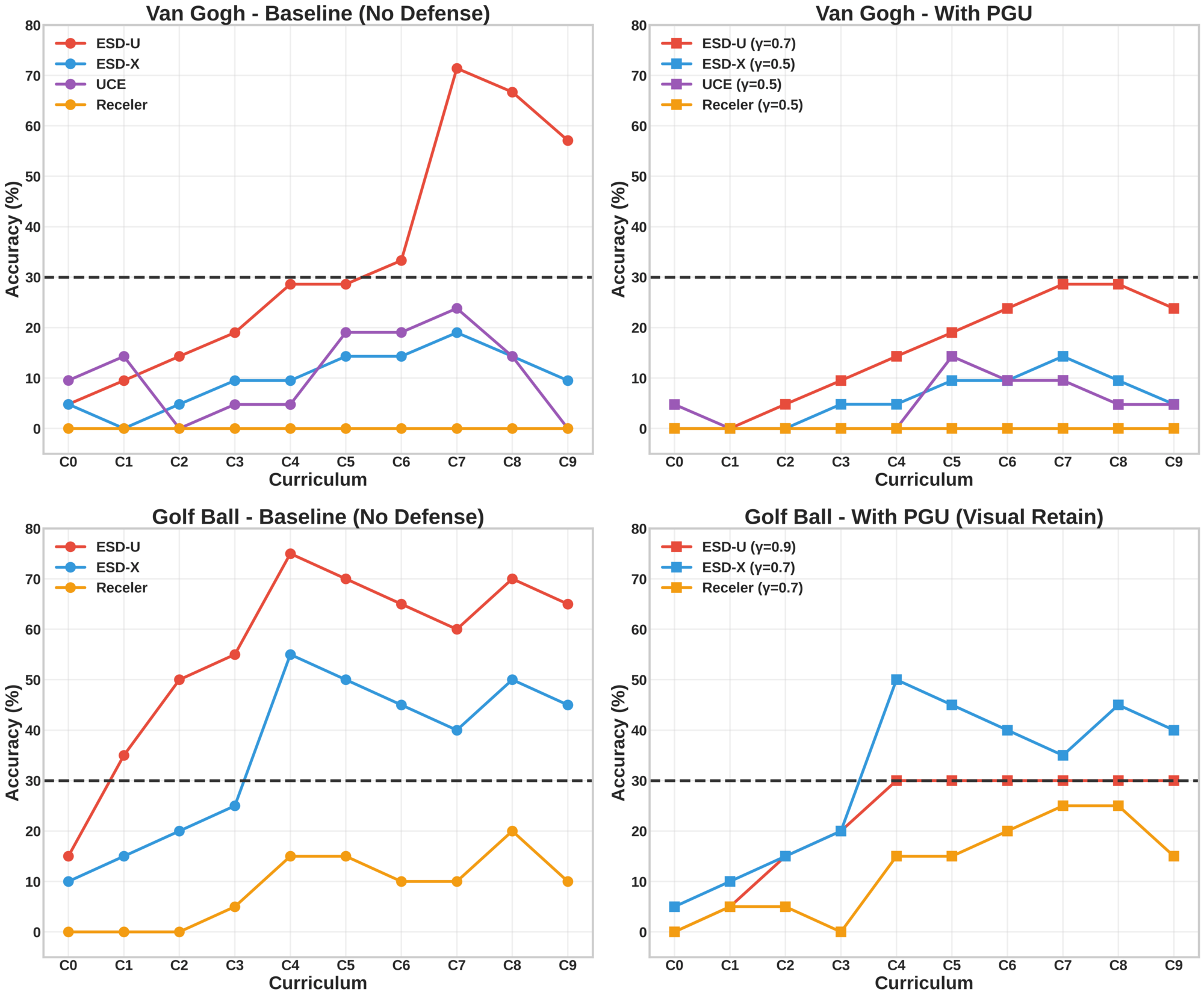}
\caption{Classifier accuracy vs.\ fine-tuning curriculum checkpoint (C0--C9) for Van Gogh
(top) and Golf Ball (bottom), showing baseline vulnerability without defense (left)
and PGU-hardened results (right). The dashed line marks the 30\% revival threshold.
PGU eliminates Van Gogh revival entirely across all methods and substantially delays
Golf Ball revival, with Receler remaining robust in both cases.}
\label{fig:all_methods}
\end{figure}

With the baseline profiles established, we now apply PGU hardening to each configuration and measure how revival points shift. Table~\ref{tab:results} and Fig.~\ref{fig:all_methods} summarize the results for all vulnerable and robust cases.

\textbf{ESD-U Van Gogh (C6$\rightarrow$None).} PGU at $\gamma=0.7$ achieves complete protection, reducing peak classifier accuracy from 71.4\% to 28.6\%, which validates the geometric hypothesis that constraining fine-tuning updates orthogonal to the retain subspace blocks gradient directions that restore erased style encodings.

\textbf{ESD-U Golf Ball (C1$\rightarrow$C4).} With visual retain concepts, PGU
achieves a 4$\times$ improvement at $\gamma=0.9$, delaying revival from C1 to C4.
Semantic retain concepts yield only a 2$\times$ improvement (C2). 

\begin{table}[H]
\centering
\caption{PGU results for Cases 1 (Vulnerable) and 2 (Robust).}
\label{tab:results}
\footnotesize
\setlength{\tabcolsep}{4pt}
\resizebox{\columnwidth}{!}{%
\begin{tabular}{@{}llccc@{}}
\toprule
\textbf{Configuration} & \textbf{Case} & \textbf{Baseline} & \textbf{Best PGU} & \textbf{Change} \\
\midrule
ESD-U VG ($\gamma$=0.7)          & Vuln. & C6 (71.4\%) & No revival (28.6\%) & Complete \\
ESD-U GB visual ($\gamma$=0.9)   & Vuln. & C1 (75.0\%) & C4 (40.0\%)         & 4$\times$ \\
ESD-U GB semantic ($\gamma$=0.9) & Vuln. & C1 (75.0\%) & C2                  & 2$\times$ \\
ESD-X GB (any $\gamma$)          & Vuln. & C4 (55.0\%) & C4 (floor)          & None \\
\midrule
ESD-X VG ($\gamma$=0.5)   & Robust & 19\% & 14\% & $-$26\% \\
Receler VG ($\gamma$=0.5) & Robust & 0\%  & 0\%  & None \\
Receler GB ($\gamma$=0.7) & Robust & 20\% & 25\% & +5 pp \\
UCE VG ($\gamma$=0.5)     & Robust & 24\% & 14\% & $-$42\% \\
\bottomrule
\end{tabular}}
\end{table}
\textbf{ESD-X Golf Ball (C4 floor).} Despite testing $\gamma\in\{0.3,0.5,0.7,0.9\}$
and both retain types, PGU consistently converges at C4 across all independent
configurations. This convergence confirms a fundamental architectural limit rather
than a configuration artifact. ESD-X breaks the text-to-visual mapping via
cross-attention modification without altering the visual knowledge stored in the convolutional
weights~\cite{paper9}; fine-tuning needs only to restore this mapping rather than
re-learn the visual geometry, requiring less gradient accumulation than PGU's
projected subspace can fully contain.

\textbf{Robust models (Case 2).} PGU consistently maintains or improves protection, confirming the ``does no harm'' property. ESD-X Van Gogh and UCE Van Gogh see peak accuracy reductions of 26\% and 42\%, respectively, increasing safety margins even for already-resistant models. For Receler, PGU preserves Van Gogh's perfect zero accuracy and produces only a minor 5 percentage point increase for Golf Ball, well below the 30\% revival threshold. The small Golf Ball shift suggests a slight interaction between PGU projection and the adapter mechanism, but remains within safe bounds.

\subsection{Case 3: Failed Erasure}
 
The UCE Golf Ball shows 80\% pre-attack accuracy, indicating that the initial unlearning itself failed. PGU produces no improvement (95\% with or without hardening). This confirms a fundamental design property: \textbf{PGU is a defense layer, not an unlearning method}. It protects erasure that has already been achieved, but cannot remove concepts that were never erased. Practitioners must verify initial erasure quality before applying PGU.
 
\subsection{Gamma Hyperparameter Ablation}

\begin{figure}[H]
\centering
\includegraphics[width=\columnwidth]{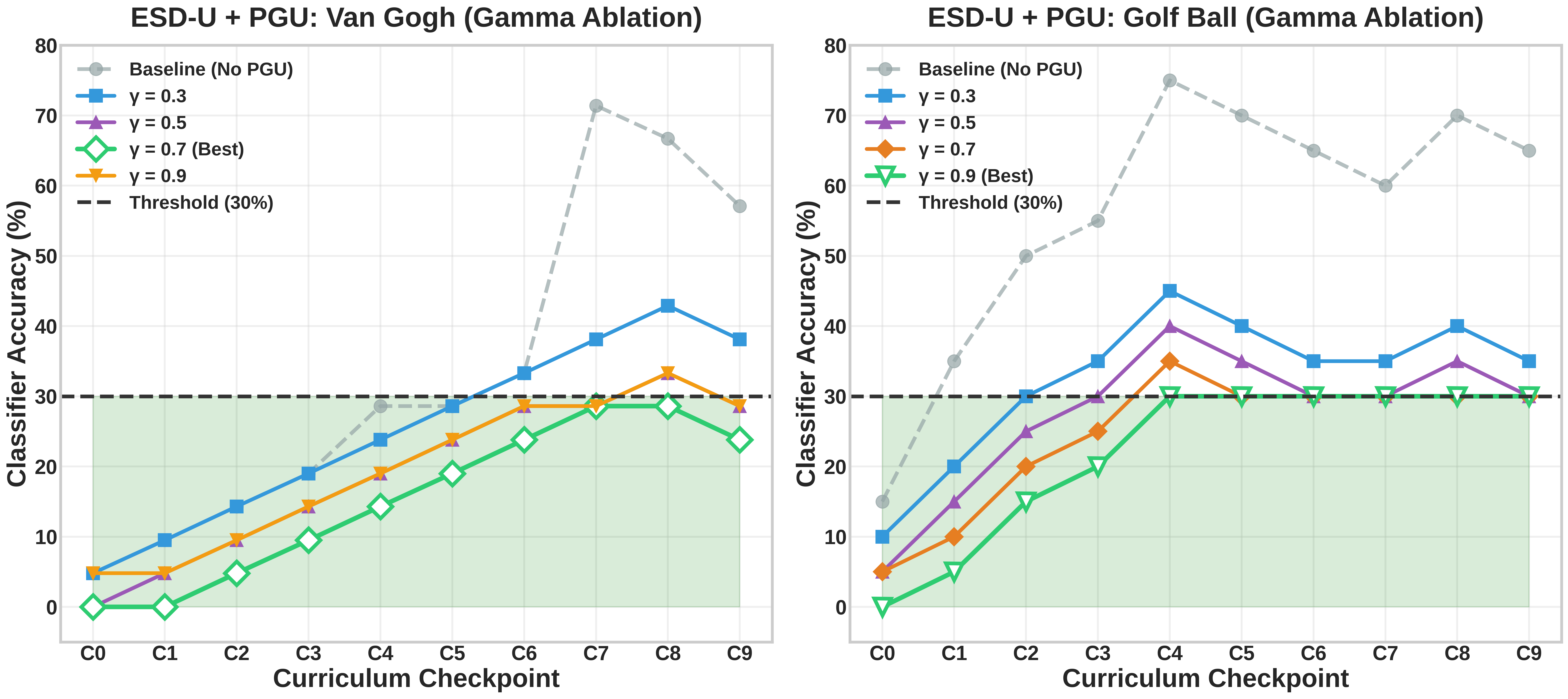}
\caption{ESD-U + PGU gamma ablation across curriculum checkpoints C0--C9. 
Left: Van Gogh style concept, where $\gamma=0.7$ (Best) keeps the classifier 
accuracy below the 30\% revival threshold throughout. Right: Golf Ball with 
visual retain set, where $\gamma=0.9$ (Best) provides the strongest 
suppression. The grey dashed line shows the undefended baseline (No PGU). 
The green shaded region indicates safe (below-threshold) classifier accuracy.}
\label{fig:gamma_ablation}
\end{figure}
\begin{table}[H]
\centering
\caption{Gamma ablation: revival points (VG\,=\,Van Gogh, GB\,=\,Golf Ball).}
\label{tab:gamma}
\footnotesize
\setlength{\tabcolsep}{9pt}
\resizebox{\columnwidth}{!}{%
\begin{tabular}{@{}llcccc@{}}
\toprule
\textbf{Method} & \textbf{Retain} & \boldmath{$\gamma$}\textbf{=0.3} & \boldmath{$\gamma$}\textbf{=0.5} & \boldmath{$\gamma$}\textbf{=0.7} & \boldmath{$\gamma$}\textbf{=0.9} \\
\midrule
ESD-U VG  & Styles   & C8   & None & None & C8 \\
ESD-U GB  & Semantic & C1   & C1   & C2   & C2 \\
ESD-U GB  & Visual   & C2   & C3   & C3   & C4 \\
ESD-X GB  & Semantic & C4   & C4   & C4   & C4 \\
ESD-X GB  & Visual   & C4   & C4   & C4   & C4 \\
\bottomrule
\end{tabular}}
\end{table}

Having established when PGU works and when it does not, we turn to how performance varies with $\gamma$, the hyperparameter controlling how much of the gradient space is reserved for retained knowledge. Diffusion models require substantially lower values than the 0.9--0.95 recommended for CNNs~\cite{paper5}. Diffusion U-Nets exhibit extensive feature sharing through skip connections, cross-attention layers that allow all concepts to share the same text-to-image conditioning mechanisms, and iterative denoising that reuses features across timesteps. This architectural entanglement means that a high $\gamma$ blocks too many gradient directions at once, impeding any meaningful fine-tuning rather than targeting revival specifically.

The concept-type dependency reflects encoding structure~\cite{7780634}: style 
representations span multiple layers through feature correlations, requiring only 
moderate $\gamma$ to disrupt revival pathways. For Van Gogh, $\gamma=0.3$ provides insufficient blocking (late revival at C8), while $\gamma\!\in\!\{0.5,0.7\}$ achieve complete protection; at $\gamma=0.9$ protection degrades back to C8, suggesting an optimal middle ground where the 
CGS is large enough to block attack directions without crowding out hardening updates, with $\gamma=0.7$ providing the lowest peak accuracy of 28.6\% (Fig.~\ref{fig:gamma_ablation}, left). Object content concentrated in higher layers benefits from more aggressive blocking: for Golf Ball, protection improves monotonically with $\gamma$, reaching optimum at 0.9 (Fig.~\ref{fig:gamma_ablation} right; Table~\ref{tab:gamma}). We recommend $\gamma=0.5$--$0.7$ for styles, $\gamma=0.7$--$0.9$ for objects, and $\gamma=0.7$ as a universal default when concept type is unknown.
 
\subsection{Retain Concept Selection}

\begin{figure}[H]
\centering
\includegraphics[width=\columnwidth]{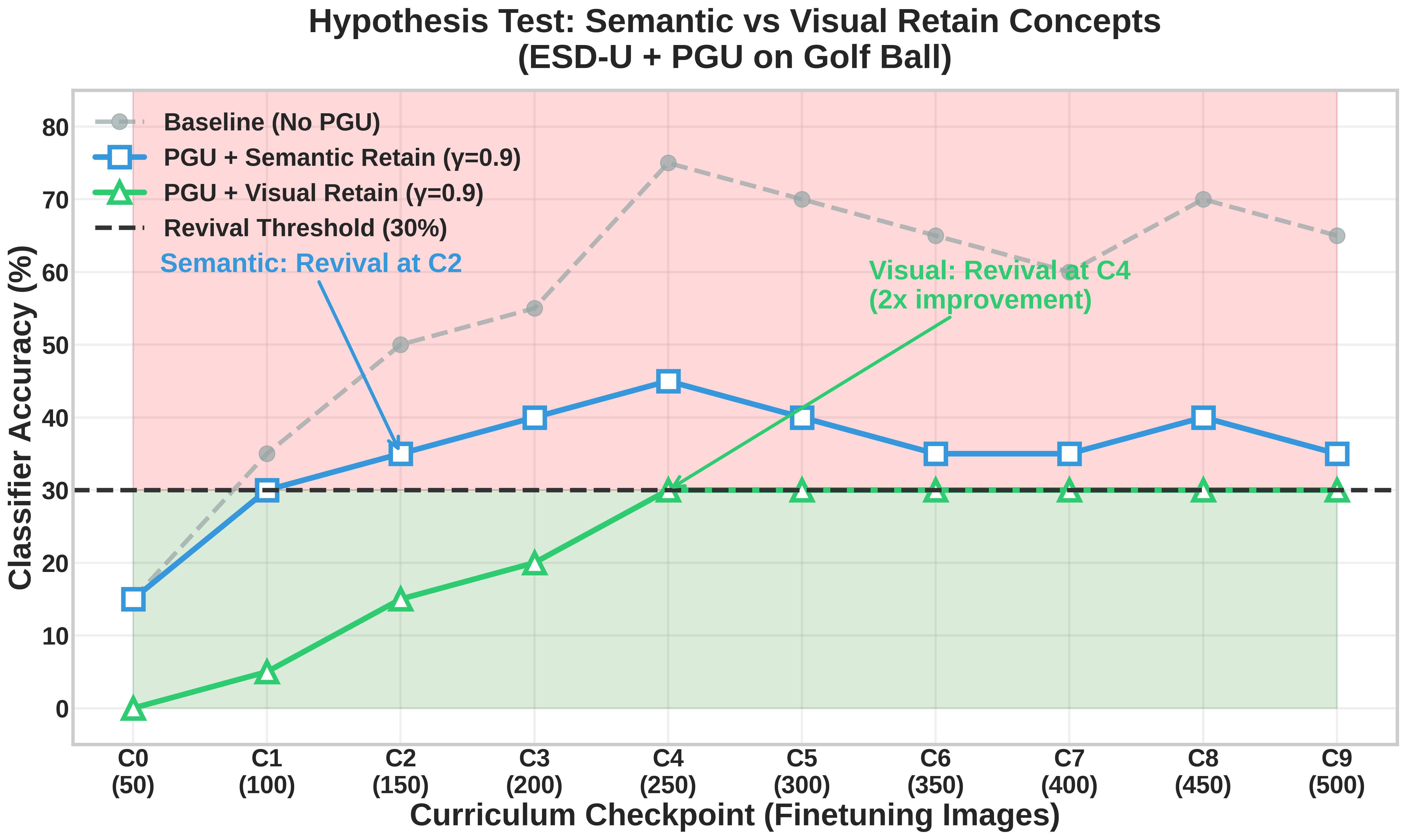}
\caption{Hypothesis test comparing semantic vs.\ visual retain concepts for 
ESD-U + PGU ($\gamma=0.9$) on Golf Ball across curriculum checkpoints C0--C9 
(50--500 fine-tuning images). Visual retain (green) delays revival to C4, 
a $2\times$ improvement over semantic retain (blue), which crosses the 30\% 
threshold at C2. The grey dashed baseline shows the undefended model. The 
red shaded region indicates revival; the green shaded region indicates 
safe (below-threshold) classifier accuracy.}
\label{fig:hypothesis_test_retain}
\end{figure}
\begin{table}[H]
\centering
\caption{Semantic vs.\ visual retain concepts for Golf Ball.}
\label{tab:retain}
\footnotesize
\setlength{\tabcolsep}{4pt}
\resizebox{\columnwidth}{!}{%
\begin{tabular}{@{}llcc@{}}
\toprule
\textbf{Method} & \textbf{Retain type (examples)} & \textbf{Best PGU} & \textbf{Gain} \\
\midrule
ESD-U & Semantic (tennis, soccer, basketball)  & C2 & 2$\times$ \\
ESD-U & Visual (ping pong, pearl, marble, egg) & C4 & 4$\times$ \\
ESD-X & Semantic                               & C4 & At floor \\
ESD-X & Visual                                 & C4 & At floor \\
\bottomrule
\end{tabular}}
\end{table}
 
Fig.~\ref{fig:hypothesis_test_retain} and Table~\ref{tab:retain} show that 
for ESD-U, visual retain concepts achieve a 2$\times$ improvement over semantic alternatives for vulnerable models. The advantage has a straightforward geometric explanation. First, the CGS is constructed from layer activations encoding visual features (shapes, textures, and spatial relationships) rather than semantic categories. Visually similar concepts (spherical, white, smooth: ping pong ball, pearl, marble, egg, moon) activate features overlapping with golf ball representations at low and mid-level layers, creating a CGS that covers gradient directions relevant to golf ball generation. Semantically similar sports balls share categorical membership but activate fundamentally different visual features, occupying distinct regions of the U-Net representation space. Second, the network encodes visual patterns through learned weights rather than explicit semantic relationships~\cite{paper7}; therefore, blocking Tennis Ball gradient directions provides limited protection against Golf Ball revival. Third, visually similar concepts create a CGS with greater geometric overlap with the target concept's gradient manifold, enabling more comprehensive blocking regardless of the specific attack direction. For ESD-X, retain concept selection has no additional effect, since the limiting factor is architectural rather than subspace coverage (C4 floor). \textit{Recommendation:} select retain concepts by visual feature overlap (shape, colour, texture) rather than semantic category.

\subsection{Comparison with Meta-Unlearning}
\label{sec:meta}
\begin{figure}[H]
\centering
\includegraphics[width=\columnwidth]
  {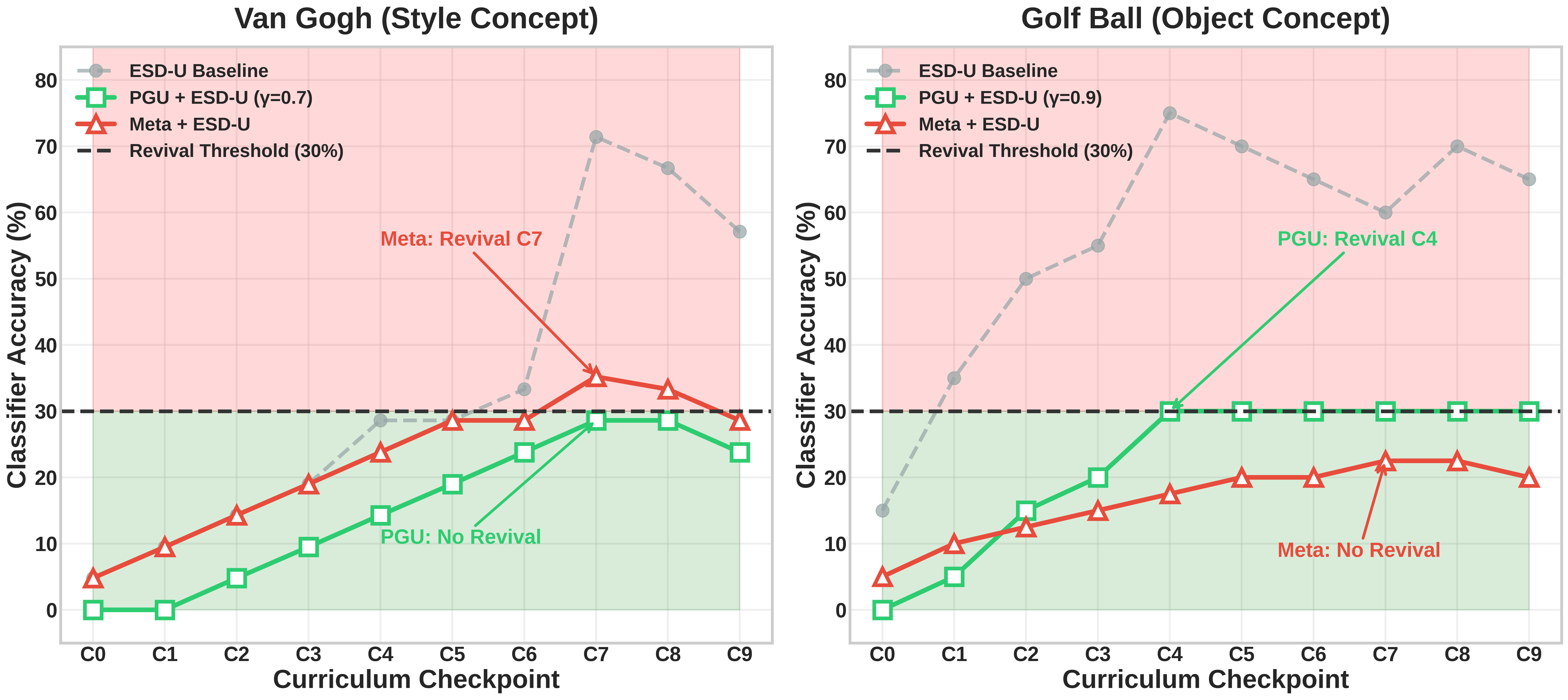}
\caption{PGU vs.\ Meta-Unlearning applied on top of ESD-U baseline across
curriculum checkpoints C0--C9. \textit{Left:} For Van Gogh (style concept),
PGU achieves no revival throughout all curricula, while Meta-Unlearning revives
at C7. \textit{Right:} For Golf Ball (object concept), Meta-Unlearning achieves
no revival while PGU reaches a protection floor at C4. The complementary
strengths align with concept encoding patterns: PGU's geometric projection
covers diverse style attack directions; Meta-Unlearning's adversarial simulation
counters the fewer, more direct object attack pathways.}
\label{fig:pgu_vs_meta}
\end{figure}

\begin{table}[H]
\centering
\caption{PGU vs.\ Meta-Unlearning~\cite{paper30} applied on top of ESD-U.}

\label{tab:meta}
\footnotesize
\setlength{\tabcolsep}{4pt}
\begin{tabular}{@{}lccc@{}}
\toprule
\textbf{Metric} & \textbf{ESD-U} & \textbf{+PGU} & \textbf{+Meta} \\
\midrule
Van Gogh revival  & C6     & \textbf{None} ($\gamma$=0.7) & C7 \\
Van Gogh max acc  & 71.4\% & \textbf{28.6\%}              & 35.2\% \\
Golf Ball revival & C1     & C4 ($\gamma$=0.9)            & \textbf{None} \\
Golf Ball max acc & 75.0\% & 40.0\%                       & \textbf{22.5\%} \\
Training time     & --     & \textbf{$\sim$6 min}         & $\sim$2 hr \\
VRAM required     & --     & \textbf{$\sim$12 GB}         & $>$40 GB \\
\bottomrule
\end{tabular}
\end{table}

Fig.~\ref{fig:pgu_vs_meta} and Table~\ref{tab:meta} reveal a clean inversion: PGU fully protects Van Gogh (no revival vs.\ Meta's C7) but hits a C4 floor for Golf Ball, where Meta-Unlearning achieves complete protection. The split follows directly from how each concept is encoded. Gatys et al.~\cite{7780634} show that artistic style is represented through correlations between filter responses across multiple network layers, a multi-scale structure that is independent of global spatial arrangement. Because style is spread across the network in this way, attack gradients arrive from many directions, each corresponding to a different aspect of the style, such as brushwork patterns, colour relationships, or textural elements. PGU's geometric projection covers entire subspaces defined by the retain styles and therefore blocks these varied attack vectors regardless of which specific direction the attack takes. Meta-Unlearning, by contrast, simulates particular attack patterns during training and cannot anticipate all the recovery pathways that such a distributed structure permits, which accounts for its bounded C7 performance.

For Golf Ball, higher-layer object representations present fewer and more direct attack pathways with predictable gradient directions. Meta-Unlearning's simulation can effectively enumerate against these limited vectors. PGU's broad geometric projection intersects these pathways but cannot fully contain the narrow gradient manifold sufficient to rebuild the text-visual mapping, resulting in the C4 floor.

In practice, concept encoding should guide the choice of defense. Distributed, multi-layer encodings, such as artistic style, are well-matched to PGU's geometric projection; compact, higher-layer encodings, such as discrete objects, are better handled by Meta-Unlearning's targeted simulation. When the concept type is uncertain beforehand, running PGU by default and adding Meta-Unlearning specifically for object targets gives the most reliable coverage.

\section{Conclusion}
This study presents the first adaptation of PGU to text-to-image diffusion models as a post-hoc defense against concept revival under fine-tuning. PGU strengthens vulnerable models, preserves already robust ones, and fails predictably when base erasure is insufficient. Compared with Meta-Unlearning, the methods are complementary: PGU handles distributed style encodings more effectively, while Meta-Unlearning is better suited to compact object encodings, and visual retain concepts consistently outperform semantic ones. The persistent C4 floor for Golf Ball and the restriction to SD v1.4 are the main limitations. Future work should pursue hybrid PGU--Meta-Unlearning strategies and validate across broader architectures.



\end{document}